\documentclass{paper}
\usepackage{spconf,amsmath,graphicx}
\usepackage{float}
\usepackage{multirow}
\usepackage{subfig}
\graphicspath{ {images/} }

\title{Ensemble of Convolutional Neural Networks for Automatic Grading of Diabetic Retinopathy and Macular Edema}
\name{Avinash Kori*, Sai Saketh Chennamsetty*, Mohammed Safwan K.P.* and  Varghese Alex* \thanks{* All authors have contributed equally.}}
\address{korivinash1@gmail.com, sakari1994@gmail.com, kpmsfn@gmail.com, varghesealex90@gmail.com}
\begin{document}
%
\maketitle
\begin{abstract}

In this manuscript, we automate the procedure of grading of diabetic retinopathy and macular edema from fundus images using an ensemble of convolutional neural networks. The availability of limited amount of labeled data to perform supervised learning was circumvented by using transfer learning approach. The models in the ensemble were pre-trained on a large dataset comprising natural images and were later fine-tuned with the limited data for the task of choice. For an image, the ensemble of classifiers generate multiple predictions, and a max-voting based approach was utilized to attain the final grade of the anomaly in the image. For the task of grading DR, on the test data (n=56), the ensemble achieved an accuracy of 83.9\%, while for the task for grading macular edema the network achieved an accuracy of 95.45\% (n=44). 

\end{abstract}
\begin{keywords}
Convolutional Neural Networks, Ensemble, Transfer learning, Diabetic Retinopathy.
\end{keywords}
\section{Introduction}
\label{sec:intro}

Diabetic Retinopathy (DR), one of the leading causes of blindness in humans, is a consequence of rupture of blood vessels in the eye and thereby leading to the discharge of blood and fluid to the surrounding tissues \cite{carrera2017automated}. Structures such as microaneurysms, hemorrhages and hard exudates are closely associated with DR and the presence of each of aforementioned anomaly determines the grade of DR in the patient. Diabetic Macular Edema (DME) is a condition that could occur at any stage of DR and is  characterized by the appearance of exudates close to the macula or retinal thickening \& thus affects the central vision of the patient \cite{al2016diabetic}. The treatment administered to subjects with DR or DME is dependent on the grade of each anomaly and thus classifying the degree of severity of DR and DME is of utmost importance. 

\par For a variety of classification and pattern recognition based tasks, Convolutional Neural Networks (CNNs) have outperformed the traditional machine learning approaches \cite{alexnet}. The superior performance of the CNNs comes at the cost of requiring millions of high quality labeled data for training the network. The presence of huge amount of labeled data in the domain of medical image analysis is extremely rare. In circumstances as stated above, a transfer learning based approach is utilized, wherein the model is first trained on a large dataset of natural images or digits and then fine-tuned for the task of choice using the limited dataset. 

\par The performance of CNNs are dependent on the architecture of the model and the connectivity pattern explored in the model. However, there exists no single architecture or connectivity which guarantees best or ideal performance. For the reason stated above, it is a common practice in deep learning to create an ensemble of classifiers with different architecture and connectivity patterns. This approach also helps in reducing the variance in the predictions made by the network. 

\par This manuscript explains our approach for the automatic grading of DR and DME from fundus images. For the task of DR grading, we make use of an ensemble of classifiers to differentiate between Normal and various variants of Non-proliferative diabetic retinopathy (NPDR) such as mild, moderate  \& severe. Additionally, an expert ensemble of classifiers were used to differentiate between Severe NPDR and proliferative diabetic retinopathy (PDR). For the task of grading DME, we make use of a one versus rest (OVR) based technique so as to mitigate the data imbalance problem in the provided dataset.

\section{Materials and Methods}

\subsection{Data}
The dataset \cite{IRID} was made available as part of the Segmentation and Grading challenge held in conjunction with the 2018 IEEE International Symposium on Biomedical Imaging. For addressing the issue of limited number of images from the ``mild" NPDR, we made use of a publicly available  mild NPDR database \cite{Additional_Data}.

\subsection{Pre-processing of Images}
The images were resized to dimension of 224 $\times$ 224 using bilinear interpolation and the intensity was normalized between the 0 and 1 using Eq. \ref{eq:norm}.

\begin{equation}
\label{eq:norm}
I_{norm} = \frac{I- min(I)}{max(I)- min(I)}
\end{equation}
Further the intensity scaled eye images were normalized to have zero mean and unit variance, Eq. \ref{eq:znorm}. The images were normalized by the same statistics (mean $\mu$ and standard deviation $\sigma$) as the ones used for pre-training the models on the Imagenet database.
\begin{equation}
\label{eq:znorm}
I_{z-norm} = \frac{I_{norm}- \mu}{\sigma}
\end{equation}

\subsection{Convolution Neural Networks}

We make use of different variants of Residual networks (ResNet) \cite{he2016identity} and densely connected networks \cite{huang2017densely}.  For the task of DR grading, 5 variants of ResNets namely ResNet-18, 34, 50, 101 \& 152 and 3 variants of DenseNets, i.e., DenseNet-121, 169 \& 201. For the task of DME grading, we use 2 variants of Resnets (Resnet-34 \& 50) and 3 variants of Densenets (Densenet-161, 169 \& 201). All the networks were pre-trained on the Imagenet dataset \cite{deng2009imagenet} and the networks used were made available by PyTorch \cite{paszke2017automatic}.

\subsection{Grading of Diabetic Retinopathy}
\subsubsection*{Training}

For the task of classifying the severity of DR in an image, we make use of 2 models; namely the primary and expert classifiers. Both the classifiers are composed of ensemble of CNNs. For training the primary classifier, the classes Severe NPDR and PDR were unified to form a new class ``S-(N)-PDR". The primary classifier classifies a fundus image as one of the 4 classes namely Normal, Mild NPDR, Moderate NPDR or S-(N)-PDR.  Each model in the ensemble was trained and validated using 70\% and 20\% of the entire training data (n=502). The models were initialized with the pre-trained weights and the parameters of networks were optimized by reducing the cross entropy loss with ADAM \cite{adam} as the optimizer. The learning rate was initialized to $10^{-3}$ and the learning rate was reduced by a factor 10\% every instance when the validation loss failed to drop. Each network was trained for 30 epochs and the model parameters that yielded the lowest validation loss were used for inference.
\par The expert classifier helps in demarcating fundus images as either one with Severe NPDR or PDR. The models in the expert classifiers were trained and validated exclusively on the images with the aforementioned classes. The training regime and all hyper-parameters such as learning rate, optimizer, learning rate decay, etc. were similar to the ones used in training the primary classifier.

\subsubsection*{Model Pruning}
Both the primary and expert classifiers are composed of 8 CNN models which varying either in terms of the depth of the network or the connectivity pattern. From the list of 8 models in the primary classifier, the model which yielded the highest number of true positives on the validation data was retained as the ``benchmark model". Furthermore, models in the ensemble those produce at least 95 percentile of number of true positives generated by the ``benchmark model" were also retained.
\par A similar approach was used to prune the models from the expert classifier ensemble also.

\subsubsection*{Testing}

\begin{figure}[H]
\centering
\includegraphics[width = 9cm]{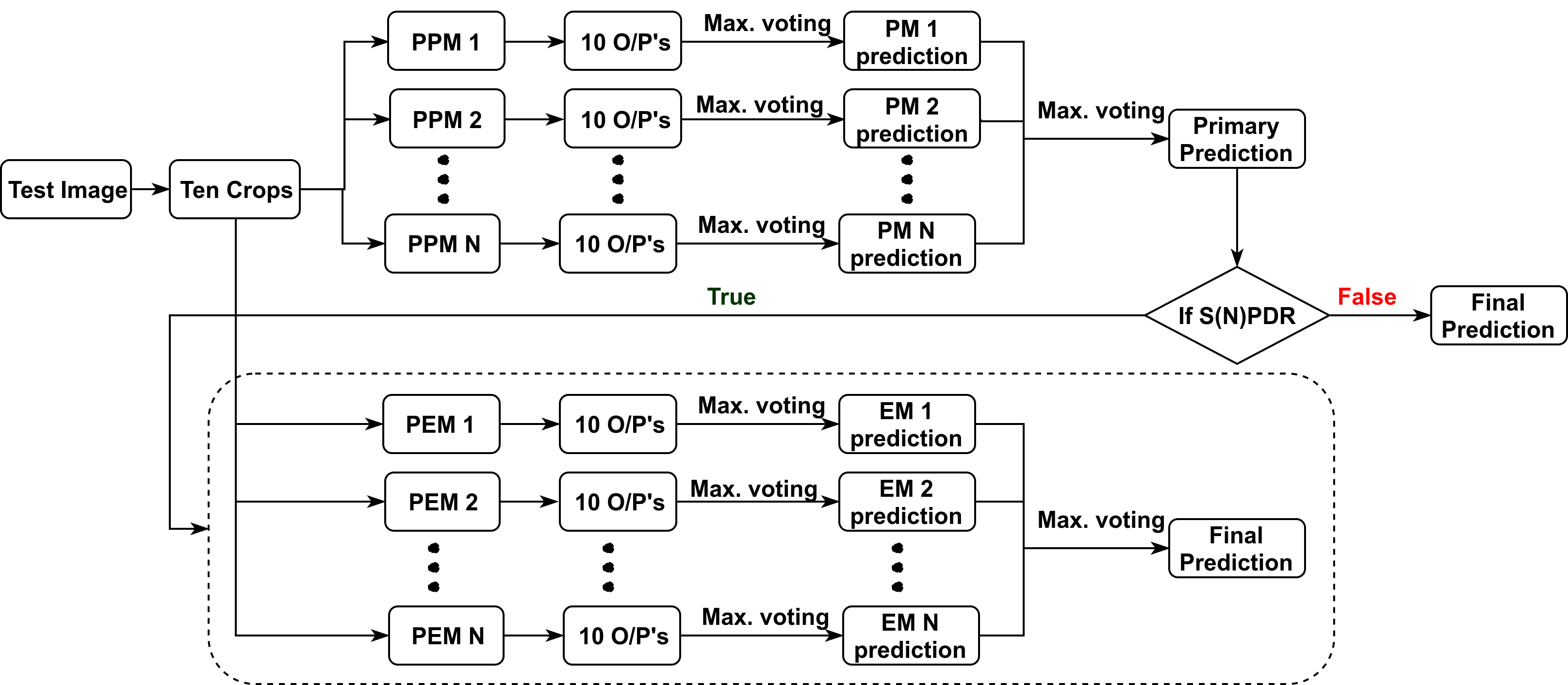}
\caption{Above figure shows exact pipeline used in our inference model. In the above figure, PPM: Pruned Primary Model, PM $X$:  prediction made by Primary Model $x$, PEM: Pruned Expert Model, EM $X$: prediction made by Expert Model $x$, O/P's: outputs}
\label{ _pipeline}
\end{figure}  

During inference, each fundus image is first resized to have a dimension of 256 $\times$ 256. From the resized images a total of 10 cropped images with a dimension of 224 $\times$ 224 were extracted. 5 cropped images were created by first cropping the image along the four corners and one along the centre of the image. Further, the image was flipped along the horizontal axis and the 5 cropped images as mentioned earlier were created to attain a total of 10 cropped images. The images are first passed through the primary classifier and optionally through the expert classifier.  

\par For a model in the pruned ensemble of primary classier, a total of 10 different variants of the test image are fed to the network. The network predicts each of the 10 variants as one of 4 classes namely ``Normal", ``Mild", ``Moderate" or ``S-(N)-PDR". The final class assigned to the test image by the model was decided by taking a maximum voting scheme.

\par The above procedure is done for all the models in the pruned-ensemble of primary classifiers. Thus for a given test image, the primary classifier returns as many predictions as the number of models in the pruned-ensemble. On the array of predictions made, a maximum voting scheme is further applied so as to classify the test image as one of the 4 classes. 

\par If the image is classified as ``S-(N)-PDR"  by the primary classifier, then the 10 cropped images are presented to pruned ensemble of the expert classifier. Each model in the ensemble classifies each cropped image as ``Severe NPDR" or ``PDR" and a maximum voting scheme is utilized  so that the model assigns a class to the image. Further, on the array of predictions made by different models in the expert classifier, a second max voting scheme is used and thereby achieve the final class assigned to the image by the expert classifier. The overall test regime is illustrated in Fig. \ref{ _pipeline}.

\subsection{Grading of Diabetic Macular Edema}
\subsubsection*{Training}
The task of DME grading is to classify each fundus image into Grade 0 (No apparent hard exudate(s)), Grade 1 (Presence of hard exudate(s) outside the radius of one disc diameter from the macula center) or Grade 2 (Presence of hard exudate(s) within the radius of one disc diameter from the macula center) depending on the presence/absence and location of hard exudate. For achieving this, we make use of 2 models based on one versus rest approach. Model 1, an ensemble of pre-trained DenseNet-161, DenseNet-169 and DenseNet-201 was trained to classify an image as class ``no apparent exudates" (Grade 0) or class ``presence of exudates" (Grade 1 or Grade 2). On the other hand, Model 2, an ensemble of DenseNet-161, Resnet-34 and Resnet-50 was trained to classify an image as ``Grade 2" DME or not. 
\par The models were trained and validated on 70\% and 20\% respectively of the entire training data. The parameters of the network were initialized with the pre-trained Imagenet data weights and the parameters were learnt by minimizing cross entropy loss with ADAM as the optimizer. The learning rate of both the networks were initialized to value of $10^{-4}$ and the learning rate was annealed step-wise with step size of 10 and the multiplicative factor of learning rate decay value of 0.9.

\subsubsection*{Testing}
\begin{figure}[]
\begin{center}
\includegraphics[width=0.5\textwidth]{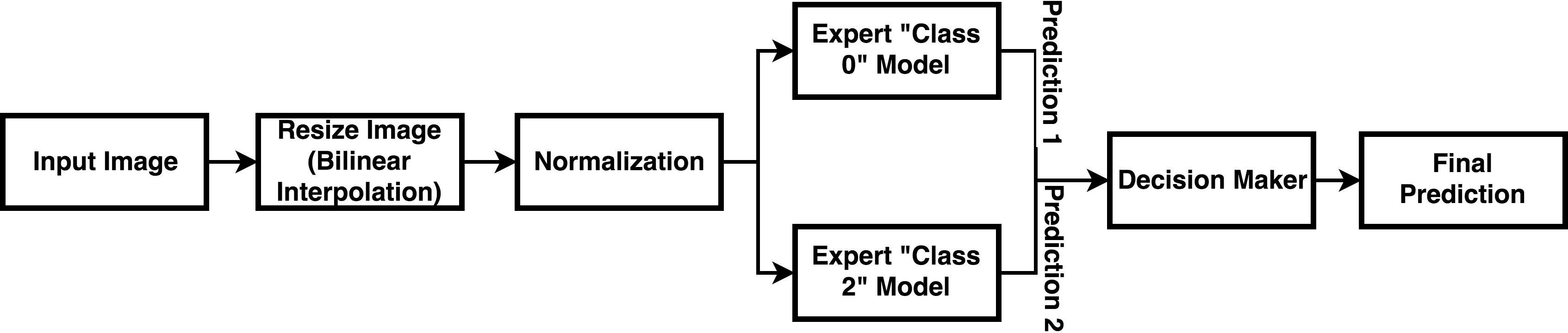}
\end{center}
\caption{Testing pipeline for Automated Diabetic Macular Edema grading.}
\label{dme_test}
\end{figure}
During inference, the images were fed to both Model 1 and Model 2 to check the presence of exudates in fundus image. The predictions of each model are then sent to a decision maker which gives out the final predictions following the rules as given in Table (\ref{Decision Maker}). The testing pipeline used for grading DME is given in Fig. \ref{dme_test}.

\begin{table}[]
\centering
\caption{DME Decision Maker}
\label{Decision Maker}
\begin{tabular}{c|c|c|c|}
\cline{2-4}
 & \begin{tabular}[c]{@{}c@{}}Model 1 \\ (Presence of \\ No Exudates)\end{tabular} & \begin{tabular}[c]{@{}c@{}}Model 2 \\ (Presence of \\ Grade 2 Exudates)\end{tabular} & \begin{tabular}[c]{@{}c@{}}Final \\ Prediction\end{tabular} \\ \hline
\multicolumn{1}{|c|}{Case 1} & True & False &  Class 0\\ \hline
\multicolumn{1}{|c|}{Case 2} & False & True & Class 2  \\ \hline
\multicolumn{1}{|c|}{Case 3} & False & False & Class 1 \\ \hline
\multicolumn{1}{|c|}{Case 4} & True & True & Class 2 \\ \hline
\end{tabular}
\end{table}

\section{Results and Discussion}
\subsection{Grading of DR}
For grading of DR, the proposed scheme was tested on a held out test data (n=56). It was observed that the technique achieved an accuracy of 84\%.  The confusion matrix achieved on the held out test data is shown in Table (\ref{my-label}).
\par When compared to the best model in the ensemble, ensembling predictions from multiple models in the ensemble produced a boost in accuracy by 1\%. Similarly we observed that rather than using all models in the ensemble, pruning or using a subset of models in the ensemble had a positive impact in the overall performance. The model pruning explored in this work helped in achieving an improvement of  1.78\% when compared to using all models in the ensemble. Providing 10 different variants of the image to the pruned ensemble during inference  was observed to produce a classification accuracy of  83.6\%, while removing the 10 crop step during inference led to a dip in classification accuracy of 6.82\%. The performance of the classifier upon using Ten Crops, pruning, etc., is given in Table (2).

\par The performance of the automated DR grading pipeline on the entire training data (n=502) is given in Table (\ref{train-label}). For classifying images as PDR in the dataset, we observe that the expert model achieves an accuracy below the expected performance, however, the expert model helps in improving the accuracy by  14\% as the ensemble of classifiers trained to demarcate all 5 classes yielded an accuracy of only 65\%. 

\begin{table}[h!]
\label{effect1}
  \begin{center}
    \caption{Effects of each step on results.}
    \begin{tabular}{|l|c|} 
    \hline
      \textbf{Method}&\textbf{Accuracy(\%)}\\
       \hline
      Without model Ensemble   & 70-74(individual model) \\
      With model Ensemble & 75\\
      \hline
      Without model Pruning  & 75 \\
      With model Pruning & 76.78\\
      \hline
      Without Ten Crops & 76.78\\
      With Ten Crops & 85.7\\
      \hline
      \textbf{Our method} &\textbf{85.7}\\
      \hline
    \end{tabular}
  \end{center}
  \label{effect}
\end{table}




\begin{table}[]
\centering
\caption{Grading of DR- Confusion Matrix on the test data. Accuracy = 85.7\%}
\label{my-label}
\scalebox{0.9}{
\begin{tabular}{|l|l|c|c|c|c|c|}
\cline{3-7}
\multicolumn{2}{c|}{}&\multicolumn{4}{c}{Prediction}&\\
\cline{3-7}
\multicolumn{2}{c|}{}& Normal & Mild & Moderate& Severe & PDR \\
\cline{1-7}
\multirow{5}{*}{\rotatebox[origin=c]{90}{Truth}}& Normal & 14 & 0 & 1 &0 &0\\
\cline{2-7}
& Mild & 2 & 10 & 0 & 0& 0\\
\cline{2-7}
& Moderate & 1 & 0 & 11& 2&0\\
\cline{2-7}
& Severe & 0 & 0 & 1 & 8 & 0\\
\cline{2-7}
& PDR & 0 & 0 & 1& 0 & 5 \\
\cline{1-7}
\end{tabular}}
\end{table}

\begin{table}[]
\centering
\caption{Grading of DR - Confusion Matrix on the training data. Accuracy = 89.4\%}
\label{train-label}
\scalebox{0.9}{
\begin{tabular}{|l|l|c|c|c|c|c|}
\cline{3-7}
\multicolumn{2}{c|}{}&\multicolumn{4}{c}{Prediction}&\\
\cline{3-7}
\multicolumn{2}{c|}{}& Normal & Mild & Moderate& Severe & PDR \\
\cline{1-7}
\multirow{5}{*}{\rotatebox[origin=c]{90}{Truth}}& Normal & 127 & 0 & 6 &1 &0\\
\cline{2-7}
& Mild & 5 & 103 & 1& 0& 0\\
\cline{2-7}
& Moderate & 4 & 0 & 128& 4&0\\
\cline{2-7}
& Severe & 0 & 0 & 8&65 &1\\
\cline{2-7}
& PDR & 0 & 0 & 5& 18 &26 \\
\cline{1-7}
\end{tabular}}
\end{table}

\subsection{Grading of DME}
On the held out test data (n=44), the proposed automated DME  grading networks achieve an accuracy of 95.45 \%. The confusion matrix of the proposed scheme on the test data is given in Table (\ref{dme-test-confusion}). 

\par On the entire training data, the proposed scheme achieves an accuracy of 96.85\%.

\begin{table}[H]
\centering
\caption{Grading of DME - Confusion Matrix on the test data. Accuracy  = 95.45\%}
\label{dme-test-confusion}
\begin{tabular}{|l|l|c|c|c|c}
\cline{3-5}
\multicolumn{2}{c|}{}&\multicolumn{3}{c|}{Prediction}&\\
\cline{3-5}
\multicolumn{2}{c|}{}& Grade 0 & Grade 1 & Grade 2\\
\cline{1-5}
\multirow{3}{*}{\rotatebox[origin=c]{90}{Truth}}& Grade 0 & 18 & 1 & 0\\
\cline{2-5}
& Grade 1 & 0 & 5 & 0\\
\cline{2-5}
& Grade 2 & 0 & 1 & 19\\
\cline{1-5}
\end{tabular}
\end{table}

\begin{table}[H]
\centering
\caption{Grading of DME - Confusion Matrix on the training data. Accuracy = 96.85\%}
\label{dme-train-confusion}
\begin{tabular}{|l|l|c|c|c|c}
\cline{3-5}
\multicolumn{2}{c|}{}&\multicolumn{3}{c|}{Prediction}&\\
\cline{3-5}
\multicolumn{2}{c|}{}& Grade 0 & Grade 1 & Grade 2\\
\cline{1-5}
\multirow{3}{*}{\rotatebox[origin=c]{90}{Truth}}& Grade 0 & 172 & 5 & 0\\
\cline{2-5}
& Grade 1 & 2 & 38 & 1\\
\cline{2-5}
& Grade 2 & 2 & 3 & 190\\
\cline{1-5}
\end{tabular}
\end{table}


\section{Conclusion}
\label{sec:pagestyle}

In this manuscript, we make use of ensemble of pre-trained classifiers for automating grading of DR and DME from fundus images. For the task of grading of DR, we observed that:
\begin{itemize}
\item Ensemble of classifiers produces better performance when compared to using a single model.
\item Selective pruning of the ensemble aids in achieving higher accuracy when compared to using all the models in the ensemble.
\item Using an expert model to differentiate severe NPDR and PDR aids in improving the performance considerably.
\item Providing 10 variants of the test image during inference aids in attaining a classification accuracy of 83.6\%.
\item On the entire training data (n =502, including additional data), the model achieved an accuracy of 89.4\%
\end{itemize}
For the task of grading of DME:
\begin{itemize}
\item The task was accomplished by training 2 ensemble of classifiers in a one vs rest fashion.
\item Similar to DR grading, ensemble of classifiers produced better performance than a single model.
\item On the entire training data, the scheme attained an accuracy of 96.85\%.
\end{itemize}

\bibliographystyle{IEEEbib}
\bibliography{refs}

	\end{document}